\documentclass{llncs}
\usepackage{makeidx}  
\usepackage{graphicx}

\usepackage[colorinlistoftodos]{todonotes}
\usepackage{listings}
\usepackage{xargs}                      
\newcommandx{\andrew}[2][1=]
{\todo[#1]{#2}}
\newcommandx{\james}[2][1=]{\todo[linecolor=red,backgroundcolor=red!25,bordercolor=red,#1]{#2}}
\newcommandx{\ken}[2][1=]{\todo[linecolor=blue,backgroundcolor=blue!25,bordercolor=blue,#1]{#2}}
\newcommandx{\kathi}[2][1=]{\todo[linecolor=OliveGreen,backgroundcolor=OliveGreen!25,bordercolor=OliveGreen,#1]{#2}}
\newcommandx{\jayashree}[2][1=]{\todo[linecolor=Plum,backgroundcolor=Plum!25,bordercolor=Plum,#1]{#2}}
\newcommandx{\jay}[2][1=]{\todo[linecolor=yellow,backgroundcolor=yellow!25,bordercolor=yellow,#1]{#2}}
\begin{document}
\mainmatter              
%


\title{DeepNeuro: an open-source deep learning toolbox for neuroimaging}
\titlerunning{DeepNeuro}  
%
\author{Andrew Beers\inst{1} \and James Brown\inst{1} \and
Ken Chang\inst{1} \and Katharina Hoebel\inst{1} \and Elizabeth Gerstner\inst{1,2} \and Bruce Rosen \inst{1} \and
Jayashree Kalpathy-Cramer\inst{1}}
\authorrunning{Beers et al.} 
%
\tocauthor{Andrew Beers and James Brown and Ken Chang and J. Peter Campbell and Susan Ostmo and Michael F. Chiang and
Jayashree Kalpathy-Cramer}

\institute{Athinoula A. Martinos Center for Biomedical Imaging,  Massachusetts General Hospital,  Charlestown, MA, USA\\
\and Department of Neuro-oncology, Massachusetts General Hosptial, Harvard Medical School, Boston MA, USA. \\
}


\maketitle              

\begin{abstract}
Translating neural networks from theory to clinical practice has unique challenges, specifically in the field of neuroimaging. In this paper, we present DeepNeuro, a deep learning framework that is best-suited to putting deep learning algorithms for neuroimaging in practical usage with a minimum of friction. We show how this framework can be used to both design and train neural network architectures, as well as modify state-of-the-art architectures in a flexible and intuitive way. We display the pre- and postprocessing functions common in the medical imaging community that DeepNeuro offers to ensure consistent performance of networks across variable users, institutions, and scanners. And we show how pipelines created in DeepNeuro can be concisely packaged into shareable Docker containers and command-line interfaces using DeepNeuro’s pipeline resources.
\end{abstract}

\section{Introduction}
\begin{figure}
\centering
\includegraphics[scale=.85,width=0.85\linewidth]{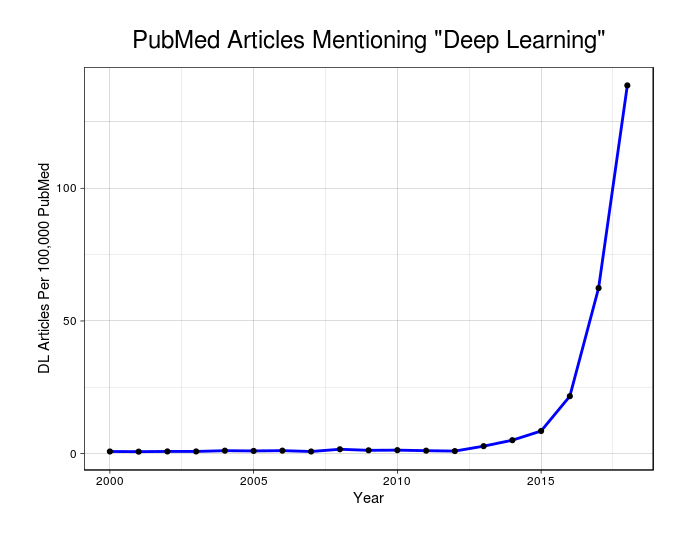}
\caption{The relative number of publications whose text or title mentions "deep learning" from 2010 to 2018 compared to all accepted publications on PubMed.}.
\label{deeplearning_growth}
\end{figure}

Deep learning is a generic term that defines an increasingly popular approach to machine learning that commonly involves learning abstract representations from datasets using learning architectures titled neural networks. With the advent of powerful graphical processing units and flexible coding frameworks, deep learning approaches have become the standard approach for computer vision tasks, and increasingly used in speech and text analysis tasks\cite{lecun2015deep,krizhevsky2012imagenet,collobert2008unified,hinton2012deep}. Naturally, deep learning is also gaining popularity in medical applications, in an era where medical imaging and written patient records are a key component of many medical workflows (Figure \ref{deeplearning_growth}). Deep learning has been successfully applied to the automated diagnosis of skin cancer, macular degeneration, diabetic retinopathy, and retinopathy of prematurity \cite{esteva2017dermatologist,lee2017deep,gulshan2016development,brown2018automated}. Specifically within the the field of neuroimaging, deep learning has been shown to have high utility for addressing pathologies as varied as Alzheimer’s disease, stroke, glioma, schizophrenia, and others \cite{liu2014early,winzeck2018isles,menze2015multimodal,bakas2017advancing,gheiratmand2017learning,chang2018residual}. In each of these problem spaces, deep learning has shown the potential for algorithms to reach accuracy and efficiency previously thought to be limited to human operators.

Despite progress at the intersection of deep learning and medical imaging, there remain  practical challenges that prevent the widespread translation of algorithms into research and clinical practice. Deep learning algorithms and neural networks are often fully-described in academic research, but the source code for implementations of these algorithms are seldom made available to other clinicians and researchers. When source code is made available, documentation is either absent, or presumes a level of familiarity with deep learning software beyond the expertise of the typical neuroimaging researcher. Even if source code for deep learning algorithms is both public and well-documented, operating system requirements and dependencies on other software packages may make their usage impractical without extensive technical support. Each of these failure points adds friction to the process of translating academic deep learning discoveries in into research practice, and further delays the point at which deep learning can be evaluated within a clinical setting.

Provided that these software challenges are overcome, the unique nature of medical imaging data produces additional barriers to the neuroscience researcher. Medical images often require numerous and highly-specialized pre- and post-processing techniques, each of which can unpredictably affect the performance of deep learning algorithms. Imaging data is often in higher resolution, as with digital pathology, or in higher dimensions, as with magnetic resonance (MR) imaging data, than traditional imaging datasets. As a result, images may need to be divided into patches, slices, or other representations before being input to a deep learning algorithm, and the specific implementation of these methods can have a significant impact on that algorithms’ performance. Post-processing techniques, particularly for segmentation algorithms, can have a significant effect on an algorithm’s practical utility in the clinic. All of these problems are even more highly elaborated in the field neuroimaging, which often has imaging sequence-specific or disease-specific processing steps for medical data. Even when such processing steps are described, a subtle change in their implementation can have significant effects on the accuracy and consistency of a deep learning algorithm. 


Other software packages have attempted to address certain aspects of these issues. NiftyNet is a software package under active development that serves as a framework and templating tool for medical imaging datasets, as well as a model repository for individual use cases \cite{gibson2018niftynet}. DLTK also serves as a framework for deep learning with medical imaging, and also has a repository for trained neural networks \cite{pawlowski2017dltk}. ModelHub.ai is an open-source, contributor-driven framework for sharing deep learning models created with any framework via Docker containers, and has an online interface for model testing. DeepInfer is a module that allows deep learning algorithms to be used within the context of 3DSlicer, a popular graphical platform for medical imaging used by researchers and clinicians alike \cite{mehrtash2017deepinfer,fedorov20123d}.

While these packages all taken together provide a strong foundation for both sharing and designing deep learning algorithms, few have the ability to do both simultaneously, and few natively provide utilities for working with data retrieved from the clinic. In this paper, we present DeepNeuro, a deep learning framework that is best-suited to putting deep learning algorithms for neuroimaging in practical usage with a minimum of friction. We will show how this framework can be used to both design and train neural network architectures, as well as modify state-of-the-art architectures in a flexible and intuitive way. We will display the pre- and postprocessing functions common in the medical imaging community that DeepNeuro offers to ensure consistent performance of networks across variable users, institutions, and scanners. And we will show how pipelines created in DeepNeuro can be concisely packaged into shareable Docker containers and command-line interfaces using DeepNeuro’s pipeline resources.

\section{Data Processing}

\begin{figure}
\centering
\includegraphics[scale=1,width=1\linewidth]{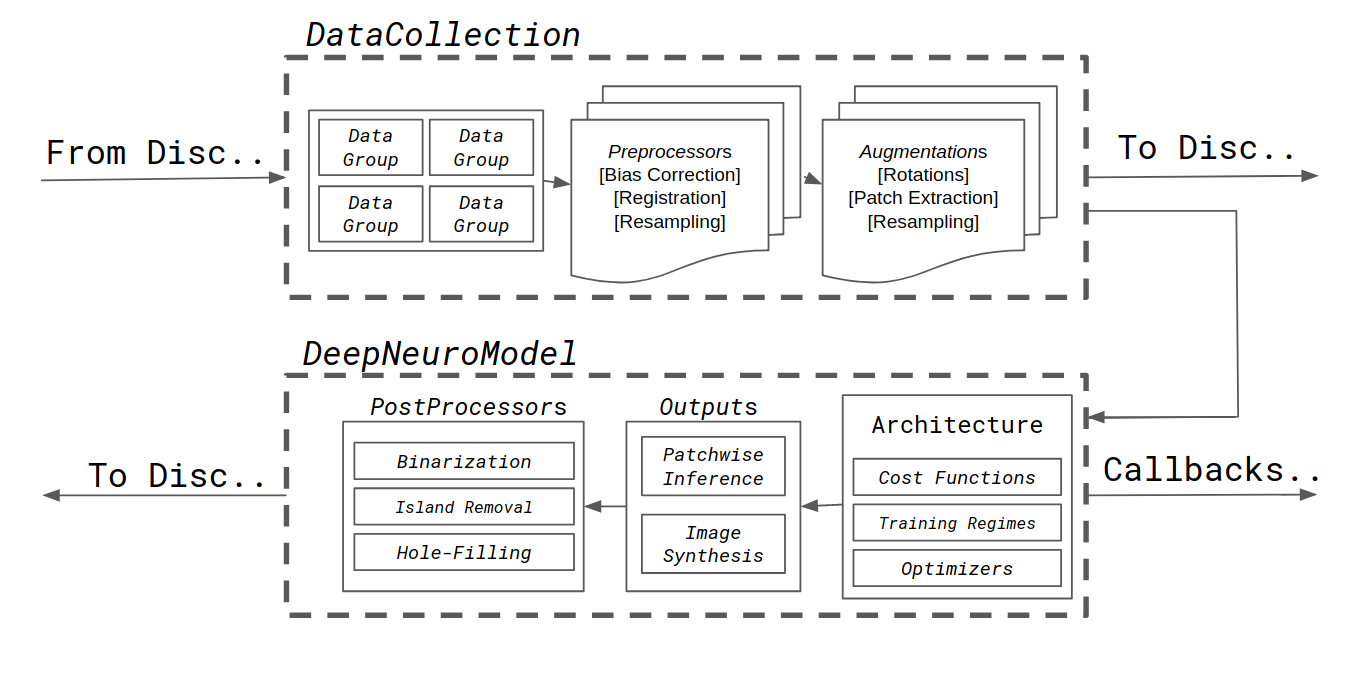}
\caption{A diagrammatic view of DeepNeuro's internal architecture. Data processing operations, such as scan normalization or data augmentation, are methods applied in the context of the \textit{DataCollection} class. A \textit{DataCollection} class can write data out, or interact with a \textit{DeepNeuroModel} class. \textit{DeepNeuroModel}s contain deep learning architectures, training regimes, and output methods such as patch reconstruction}.
\label{box_model}
\end{figure}

\subsection{Data loading}
The DeepNeuro scripting language is centered on the \textit{DataCollection} class. \textit{DataCollections} are Python objects that stores listed information about any data to be input into later processing steps (Figure \ref{box_model}). Each dataset is conceptualized as a series of “cases”, or individual groups of images and metadata upon which other DeepNeuro methods act. \textit{DataCollections} have flexible data loaders to associate and stack different medical imaging inputs (e.g. sequences, modalities) into stacked NumPy arrays from regular folder structures or .csv lists to file paths \cite{walt2011numpy}. \textit{DataCollections} store both information about the location of data on disk, derived attributes of the data itself -- e.g. data shape, dimension, and intensity range, and metadata if provided in the original disk data format.

\textit{DataCollections} list separate Python objects called \textit{DataGroups}. \textit{DataGroups} are subsets of a given patient case that may be input in different sections of a model. The most typical \textit{DataGroups} designations are \textit{input\_data}, i.e. the input node of a neural network, and \textit{ground\_truth} data, against which the cost function of neural network can be evaluated. However, other groups can be specified for more complex architectures, such as networks that contain inputs at multiple nodes. \textit{DataCollections} can sample data such that the same data from each \textit{DataGroup} is sampled in each batch, or such that \textit{DataGroups} are sampled randomly, as in the case of unpaired generative adversarial networks.

\textit{DataCollections} have the ability to save their inputs into HDF5 file format, and load from HDF5 file format without changes in data organization. \textit{DataCollections} utilize lazy loading, only loading data to generate attributes upon request.

\begin{figure}
\centering
\includegraphics[scale=1,width=1\linewidth]{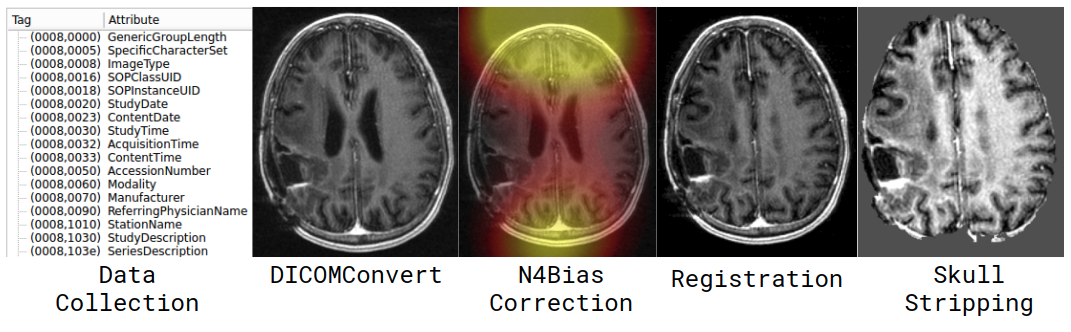}
\caption{An example of a typical pre-processing script for a neuroimaging pipeline. Data is converted from a directory of DICOM files into an internal representation in NumPy, corrected for intensity bias using 3DSlicer's N4ITKBiasCorrection tool, co-registered across sequences using 3DSlicer's BRAINSFit tool, and skull-stripped using a neural network trained in DeepNeuro.}.
\label{preprocessing_fig}
\end{figure}

\subsection{Preprocessors and Postprocessors}
\subsubsection{Preprocessing}
Standardized preprocessing methods are essential to deep learning pipelines that operate on medical images, as slight differences in preprocessing methods can lead to catastrophic prediction failures. DeepNeuro allows the user to preprocess data before inference using the Python object \textit{Preprocessor}. Transformations applied in \textit{Preprocessor} objects can be pure Python implementations, inference via neural networks, or links to outside programs such as 3DSlicer or ANTs \cite{avants2014insight}. These transformations can be applied to data held either in memory or loaded from a provided filepath, and can be returned as either Numpy arrays or stored back to disc. Preprocessors can be concatenated sequentially into preprocessing pipelines, or applied selectively to certain data objects and not others. Current \textit{Preprocessor} objects available include 3D image registration (3DSlicer), 3D image resampling (3DSlicer), N4 Bias Correction (3DSlicer, ANTs), and skull stripping using a model trained with DeepNeuro (Figure \ref{preprocessing_fig}).

\subsubsection{Postprocessing}
\textit{Postprocessor} objects share the same structure and capabilities as \textit{Preprocessor} objects, but are applied to \textit{DeepNeuroModel} objects (described below) instead of \textit{DataCollections}. \textit{Postprocessors} are used to apply transformations to data that has been generated by a model. \textit{Postprocessor} transformations include island-removal and hole-filling for binary outputs, and binarization for scalar outputs.

\subsection{Data Augmentation}
Data augmentation is used to functionally increase the size of datasets fed into machine learning models via spatial or contrast-based data transformations. Data augmentation is especially important in medical imaging, as medical image datasets tend to be far smaller than datasets of natural images. DeepNeuro addresses this need with \textit{Augmentation} objects, which can be assigned to \textit{DataCollection} objects. Each \textit{Augmentation} object specifies a data augmentation to be applied to input data either before writing to HDF5, or lazily during training. \textit{Augmentations} can be applied for all \textit{DataGroups}, or selectively for specific \textit{DataGroups}. Augmentations are specified in a sequential fashion, and a recursive method is used to concatenate transformations on a single data input. Data can be augmented before saving to HDF5 format, and augmented data is randomly shuffled before sampling in batches. 

\begin{figure}
\centering
\includegraphics[scale=1,width=1\linewidth]{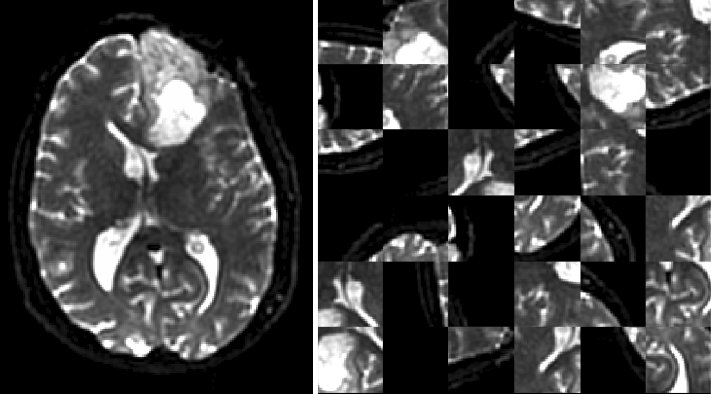}
\caption{A sample slice of a magnetic resonance image of the brain, and some augmented patches at size 128x128 pixels created by DeepNeuro's \textit{PatchAugmentation} and \textit{Flip\_Rotate\_2D} augmentation classes.}
\label{augmentation_fig}
\end{figure}

Current \textit{Augmentation} objects available include 2D and 3D flips and rotations, intensity scaling and shifting, 3D patch extraction, channel-wise dropout, and nearest-neighbor downsampling. Patch extraction can be performed to preferentially select patches that match certain criteria, such as being near a tissue of interest (Figure \ref{augmentation_fig}).

\subsection{Additional Utilities}

\subsubsection{Medical Image Data Formats}
Data loading features for NiFTI, NRRD, and DICOM files are also included as standalone functions, as well as data saving functions for both NiFTI and DICOM Segmentation Objects (DSO) via the external package dcmqi \cite{herz2017dcmqi}.

\section{Model Design}

\subsection{DeepNeuroModels}

Language abstraction is common in programming frameworks for neural networks. Keras, one of the most popular frameworks, is an abstraction that can be run with several different backends, including TensorFlow, Theano, and MXNet \cite{abadi2016tensorflow,bergstra2010theano,chen2015mxnet}. These, in turn, are Python abstractions over lower-level languages. Still other families exist outside of the Keras framework, such as the popular academic deep learning framework PyTorch \cite{paszke2017automatic}. Inadvertently, this wide variety of languages has created some difficulty in making models open-source, as researchers split between TensorFlow, PyTorch, or another framework may have difficulty readily implementing each others’ code.

The \textit{DeepNeuroModel} is another level of abstraction designed to address this problem within the practical context of DeepNeuro pipelines. \textit{DeepNeuroModels} are objects that take \textit{DataCollections} as inputs, and can process the data stored in that object via a set of standardized functions shared across deep learning frameworks. These include model training, model saving, generating model callbacks for training loss and other features, and performing inference. This minimum set of functions is presently implemented via subclasses for Keras and Tensorflow, and can be called agnostically of the original framework from DeepNeuro’s interface.

\subsubsection{Model Customization}
When possible, all models are constructed to be implemented in 2D or 3D. The models available in DeepNeuro have several options for parameter customization, and thus hyperparameter optimization. These parameters include model depth, filter number, dropout ratios, batch normalization, activation type, cost functions, kernel size, and stride size. Models can take advantage of custom cost functions more common in medical imaging, such as the soft-dice cost function and the Wasserstein gradient penalty. Parameter specific to training regimes can also be specified, including batch sizes, learning rates, and optimizers.

\subsubsection{Model Implementations}

For segmentation applications, we have implemented the U-Net architecture in both 2-D and 3-D \cite{ronneberger2015u,cciccek20163d}. For image synthesis applications, we have implemented both a traditional generative adversarial network architecture (GAN) and the progressively growing GAN architecture proposed by Karras et. al \cite{karras2017progressive}. 

To improve performance of the U-Net, we incorporate state-of-the-art components that have improved neural network architectures for classification tasks, namely inception modules, residual connections, dense connections, and squeeze-and-excitation modules. \cite{szegedy2015going,he2016deep,huang2017densely,hu2017squeeze} Inception modules have multiple pathways with different convolution filter sizes. Residual connections are “shortcut” connections that allow for bypass of convolution layers. Dense connections allow feature maps from every convolutional layer to be carried forward. Squeeze-and-excitation modules allow learning of relationships between different feature maps. We also allow for easy modification of the neural network architectures for changing the ordering of batch normalization, convolutional, and activation layers. \cite{he2016identity}


\subsubsection{Inference and Outputs}

DeepNeuro includes \textit{Output} objects, which are processes that can be generated from \textit{DeepNeuroModel} objects. The primary subclass is \textit{Inference}, which computes the output of a \textit{Model} after being fed input data. Inference itself is subclassed, for example with \textit{ModelPatchesInference} for patch-based models. \textit{ModelPatchesInference} provides options such as the degree of patch overlap (i.e., how many patches to extract and average) and whether to pad the input data to ensure full coverage.

\section{Pipeline Distribution}

DeepNeuro is designed to be packaged into human-readable modules, and then distributed through Docker containers. The \textit{pipelines} module of DeepNeuro contains several example pipelines (detailed in “Sample Applications”) that give a simple framework for building DeepNeuro training and inference modules.

DeepNeuro also contains utilites to package Docker containers into command line utilities, with examples in the \textit{pipelines} module. These command-line modules process datasets individually, are data format agnostic, and let the user specify the number of preprocessing steps required for their particular dataset. DeepNeuro Docker containers are based on the \textit{nvidia-docker} runtime in order to take advantage of NVIDIA GPUs for neural network inference. Docker containers exist for each module, containing the base DeepNeuro container and the models necessary to run that particular module.

Models are stored outside of DeepNeuro in a cloud-based data management system, and either come pre-downloaded via Docker containers, or can be downloaded via the load modules in DeepNeuro. Models are stored within the DeepNeuro library, and can be deleted via the load module in the scenario where the user wishes to train a newly initialized model.

\section{Sample Modules}

\subsection{Brain Extraction}

\begin{figure}
\centering
\includegraphics[scale=.5,width=.5\linewidth]{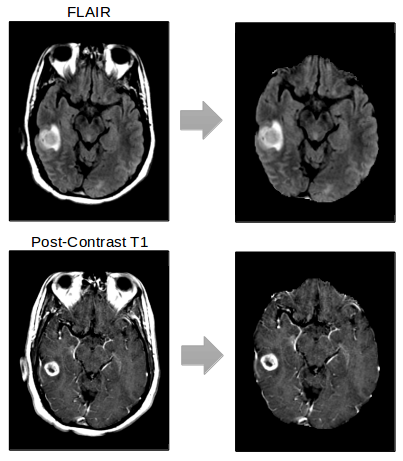}
\caption{Example of DeepNeuro's skull-stripping module on a patient with a brain tumor. Fluid-attenuated inversion recovery sequences and post-contrast T1 sequences are fed into DeepNeuro's skull-stripping module, and a binary mask of the brain is provided as output.}
\label{skullstrip_module}
\end{figure}

Brain extraction or ``skull-stripping'' is a common image preprocessing step that is essential for many neuroimaging applications, including cortical parcellation and surface reconstruction. \cite{kleesiek2016deep} Many effective, computationally-efficient tools exist for performing brain extraction. \cite{jenkinson2012fsl,segonne2004hybrid,cox1996afni,shattuck2002brainsuite,iglesias2011robust} However, their generalizability is limited in the presence of varying acquisition parameters or abnormal pathology, such as tumors. Without manual correction, poor brain extraction can introduce errors in downstream analysis. Using DeepNeuro’s U-Net architecture with the following parameters, we trained on a dataset of 30 glioma patients from a multi-institutional cohort, with manually-segmented brain masks (Figure \ref{skullstrip_module}). This patient cohort contained both pre-operative and post-operative patients.

\begin{lstlisting}[language=Python]
    from deepneuro.models.unet import UNet
    model_parameters = {'input_shape': (64, 64, 8, 2),
                    'pool_size': (2, 2, 1), 
                    'kernel_size': (3, 3, 3),
                    'dropout': 0, 
                    'batch_norm': True, 
                    'initial_learning_rate': 0.00001, 
                    'cost_function': 'soft_dice',
                    'num_outputs': 1, 
                    'activation': 'relu',
                    'padding': 'same',
                    'backend': 'keras',
                    'depth': 4,
                    'max_filter': 512,
                    'downsize_filters_factor': 1,}
    unet_model = UNet(**model_parameters)
\end{lstlisting}

\subsection{Glioblastoma Segmentation}

\begin{figure}
\centering
\includegraphics[scale=.9,width=.9\linewidth]{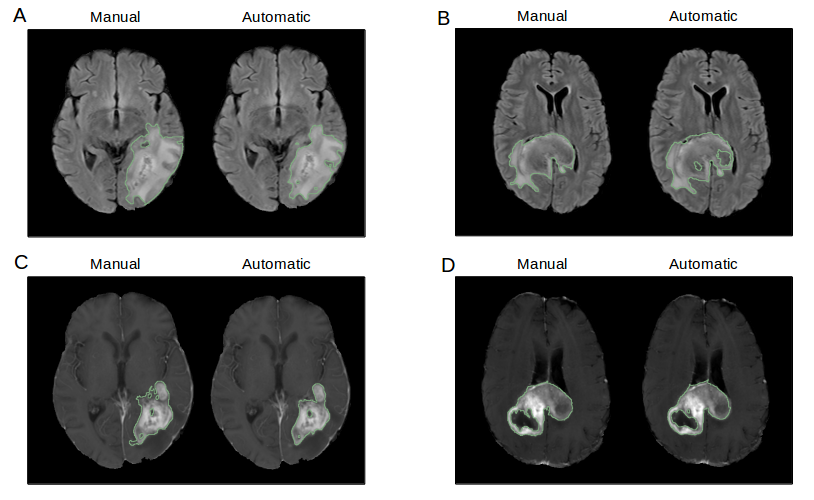}
\caption{DeepNeuro segmentations displayed compared with sample manual segmentation performed by an expert neurooncologist with 5+ years of experience. A) and B) show binary whole tumor (edema, enhancing tumor, necrosis, and non-enhancing tumor) masks overlaid on a FLAIR sequence, while C) and D) show binary enhancing tumor masks overlaid on post-contrast T1 sequences.}
\label{segmentation_module}
\end{figure}

	Pathological volume monitoring, and thus pathological tissue segmentation, is essential for assessment of treatment response and prognosis in glioblastoma treatment. \cite{brasil2009low,iliadis2012volumetric} Furthermore, volumes and imaging features derived from delineated tumor regions can be used for downstream prediction of molecular biomarkers, treatment response, progression, and survival.\cite{zhang2016multimodal,grossmann2017quantitative,smits2017imaging,chang2016multimodal}. Unfortunately, manual delineation of tumor boundaries can be challenging and subject to inter- and intra-rater variability, resulting in low reproducibility even among expert radiologists and oncologists. Additionally, it is a laborious task especially for high-resolution scans which can have numerous image slices. This diverts clinicians’ time away from other clinical and research tasks, as well as other patients. There are two tumor regions that are of key interest to the clinician. The first is the whole tumor, which consists of edematous tissue, non-enhancing, enhancing tumor, and necrosis. This is best seen on the T2 FLAIR sequence and represents the total tumor burden. The second is contrast-enhancing tumor, which represents regions of breakdown of the blood brain barrier. \cite{dubois2014gliomas}
    
	The Glioblastoma Segmentation module uses two 3D U-Net architectures. The first creates a binary labelmap of a region-of-interest defined as whole tumor. The output of this network is fed as an additional channel into a second network, which predicts a binary labelmap of enhancing tumor alone (Figure \ref{segmentation_module}). Both networks take in 32x32x32mm patches extracted from FLAIR, pre-contrast T1, and post-contrast T1 patient MR sequences, stacked channel-wise.
    
	Both U-Nets have a depth of four max-pooling layers, with two convolutional layers between each pooling layer, leading to a U-Net architecture with 18 convolutional layers. The network is trained on the BRATS 2017 dataset as well as a clinical trial patient cohort from the Massachusetts General Hospital \cite{menze2015multimodal,bakas2017advancing,bakas2017segmentationgbm,bakas2017segmentationlgg}.

\begin{lstlisting}[language=Python]
    from deepneuro.models.unet import UNet
    model_parameters = {'input_shape': (32, 32, 32, 3),
                    'pool_size': (2, 2, 2), 
                    'kernel_size': (5, 5, 5), 
                    'dropout': 0, 
                    'batch_norm': True, 
                    'initial_learning_rate': 0.0001, 
                    'cost_function': 'soft_dice',
                    'num_outputs': 1, 
                    'activation': 'relu',
                    'padding': 'same', 
                    'backend': 'keras',
                    'depth': 4,
                    'downsize_filters_factor': 1,
                    'max_filter': 256}
    unet_model = UNet(**model_parameters)
\end{lstlisting}

\section{Future Directions}

We present a Python package and model distribution system entitled DeepNeuro. It is a framework for generating and training neural network architectures across multiple programming backends, an all-in-one data preprocessing tool for neuroimaging, and a templating and distribution system for end-to-end deep learning algorithms in neuroimaging.

We will continue to add features to DeepNeuro, and encourage contributions from the users of DeepNeuro in the form of both features and additional modules. We particularly anticipate expanding DeepNeuro’s support for PyTorch, expanding the breadth of model templates available for users to train on, and creating a GUI interface for creating DeepNeuro pipelines for those without scripting proficiency. We also plan to expand the subclasses of \textit{DeepNeuroModel} to include models created with MXNet and PyTorch, to facilitate the rapid development of pipelines in these languages.

\section{Acknowledgements}

The Center for Clinical Data Science at Massachusetts General Hospital and the Brigham and Woman's Hospital provided technical and hardware support for the development of DeepNeuro, including access to high-powered graphical processing units.

%
%

\bibliographystyle{splncs}
\bibliography{thebib}




\end{document}